\documentclass[runningheads]{llncs}
\usepackage[T1]{fontenc}
\usepackage{graphicx}
\usepackage{times}
\usepackage{latexsym}
\usepackage{graphicx} 
\usepackage{amsmath}
\usepackage{amsfonts}
\usepackage{wrapfig}
\usepackage{subcaption}

\usepackage{url}
\usepackage{hyperref}
\usepackage{enumitem}

\usepackage{xcolor}
\hypersetup{
    colorlinks,
    linkcolor={red!50!black},
    citecolor={blue!50!black},
    urlcolor={blue!80!black}
}

\begin{document}
\title{Investigating OCR-Sensitive Neurons to \\ Improve Entity Recognition in Historical Documents}
%
\titlerunning{Investigating OCR-Sensitive Neurons to Improve Historical NER}
%

\author{Emanuela Boros\orcidID{0000-0001-6299-9452} \and
Maud Ehrmann\orcidID{0000-0001-9900-2193}}

%
\authorrunning{E. Boros and M. Ehrmann}
%
\institute{Digital Humanities Laboratory, EPFL, Lausanne, Switzerland \\
\email{emanuela.boros@epfl.ch, maud.ehrmann@epfl.ch}\\
}
%
\maketitle              
\begin{abstract}

This paper investigates the presence of OCR-sensitive neurons within the Transformer architecture and their influence on named entity recognition (NER) performance on historical documents. By analysing neuron activation patterns in response to clean and noisy text inputs, we identify and then neutralise OCR-sensitive neurons to improve model performance. Based on two open access large language models (Llama2 and Mistral), experiments demonstrate the existence of OCR-sensitive regions and show improvements in NER performance on historical newspapers and classical commentaries, highlighting the potential of targeted neuron modulation to improve models' performance on noisy text.\footnote{This version of the contribution has been accepted for publication at ICADL 2024, after peer review but is not the version of record and does not reflect post-acceptance improvements, or any corrections.}


\end{abstract}

\section{Introduction}

\paragraph{\bf Context and Motivation.}
The ability to accurately process and extract information from historical documents is essential for the computational analysis and study of digitised archives. 
Over the last decade, numerous projects have emerged  to mine the rich information contained in unstructured texts obtained via optical character recognition (OCR),  aiming notably to support exploration and discovery~\cite{doucet_newseye_2020,ehrmann_language_2020}, enable computational analysis~\cite{smith_computational_2015,mcgillivray_challenges_2020}, and improve cultural heritage data availability and curation~\cite{rehm_qurator_2020,padilla_responsible_2020,candela_reusing_2022}. While these efforts are promising and to a large extent already successful, they face a recurring challenge, namely the variable and often mediocre quality of OCR transcriptions. Misrecognised characters and words, sometimes affecting entire sections of text, severely impact the performance of downstream processes~\cite{vanstrien_assessing_2020}, including language modelling~\cite{todorov_assessment_2022a} and named entity processing \cite{linharespontes_impact_2019a,hamdi_assessing_2020a,ehrmann_overview_2022,ehrmann_named_2023a}. Whatever the task, the document type and the language, OCR noise is a pervasive issue that undermines the robustness of deep neural language models, which are affected by even slightly perturbed input~\cite{ehrmann_extended_2020b,moradi_evaluating_2021}.

Various approaches have been and are being developed to mitigate OCR noise. These include improving transcription accuracy~\cite{neudecker_ocrd_2019}, devising post-correction techniques~\cite{huynh_when_2020a,boros_postcorrection_2024} and increasing the resilience of models to corrupted input. In addressing the latter, approaches have focused on adapting either the model architecture or the training data. For NER, Boros \textit{et al.}~\cite{boros_alleviating_2020} proposed enhancing a fine-tuned BERT model~\cite{devlin_bert_2019} by stacking two additional Transformer blocks on top of it~\cite{vaswani_attention_2017a}. This approach effectively improved the model's handling of OCR noise, achieving top performance in the HIPE-2020 campaign~\cite{ehrmann_extended_2020b}. Schweter \textit{et al.}~\cite{schweter_hmbert_2022a} adopted a data-centric approach for the same task by pre-training a language model on curated historical data before fine-tuning it. Although this approach came close to the stacked Transformer method, it was no better. Similarly, Manjavacas \textit{et al.}~\cite{manjavacas_adapting_2022} investigated several methods of pre-training on historical data and concluded that matching pre-training with target data is beneficial. Despite these efforts, the challenge of improving model robustness to OCR noise remains largely unresolved. As a complement to these data- and architecture-centric approaches, this paper proposes to investigate the sensitivity of the internal components of models to OCR noise.

\vspace{0.1cm}
\noindent \textbf{\bf Background.}
Motivated by the opacity of deep neural networks, a growing body of research focuses on interpreting and analysing models to understand what knowledge they learn and how they use it to make decisions. A first line of research aims to determine what information is captured where in a network by analysing the learned (vector) representations. Besides visualisation~\cite{hupkes_visualisation_2018}, the most common method is to assess whether internal representations (e.g. word or sentence embeddings, gate activations, attention weights) are associated with a given property by training a classifier on these (frozen) representations to predict the said property. Such diagnostic classifiers have been used to probe a variety of morphosyntactic and semantic properties~\cite{adi_finegrained_2017,conneau_what_2018a,belinkov_linguistic_2020,sajjad_analyzing_2022a}, as well as non-linguistic ones~\cite{gurnee_language_2024a} (see also \cite{belinkov_analysis_2019} for an overview and \cite{belinkov_probing_2022b} for a critique of this method).

Complementary to this representation analysis, recent research focuses on the role of individual neurons or groups of neurons. The aims are to understand how properties or concepts are learned by neurons, to identify specific abilities of particular neurons, and to determine which ones are involved in predictions and whether they can be inhibited~\cite{sajjad_neuronlevel_2022}. Neuron-level analyses often involve studying activation patterns in response to different inputs to identify neurons that react to specific input features or tasks. Main methods include visualisation of neuron activations~\cite{erhan_visualizing_2009,karpathy_visualizing_2015}, statistics of activation values in response to a set of inputs~\cite{na_discovery_2019}, and neuron probing~\cite{dalvi_what_2019}. This allows, among others, the identification of neurons that learn specific  properties~\cite{bau_identifying_2018,na_discovery_2019,durrani_analyzing_2020,wang_finding_2022}, factual knowledge~\cite{dai_knowledge_2022}, or are language  specific~\cite{tang_languagespecific_2024a}. It also supports the discovery of redundant units, thereby aiding feature selection and model distillation~\cite{dalvi_analyzing_2020}, and enables control of neurons by modifying their activations~\cite{bau_identifying_2018}.

\vspace{0.1cm}
\noindent \textbf{\bf Objective.}
With this in mind, this study aims to 1) assess whether some model components, specifically layers and neurons, are sensitive to OCR noise, and 2) determine whether these components can be controlled to reduce their adverse effect on NER performance in historical documents. 
We address these questions by measuring differences in the activation of model components in response to clean and noisy inputs, seeking to identify those that consistently react to disturbed text (Section \ref{sec:sensitive_components}). We then leverage our findings to attempt to neutralise OCR-sensitive neurons in the use case of detecting named entities in historical documents (Section \ref{sec:downstream-task}). 


\section{Discovering OCR-sensitive Layers and Neurons}
\label{sec:sensitive_components}

Our approach is twofold: first, we try to identify regions of the network affected by noise by focusing on whole representations. 
Second, we closely examine the behaviour of individual neurons. Both experiments share the same setup, which we present first.

\subsection{Experimental Setup}
\vspace{-0.4cm}
\vspace{0.1cm}
\noindent \textbf{\bf Models.}
This study focuses on the decoder-only Transformer architecture and is based on two pre-trained large language models (LLM), Llama2 and Mistral. Released by Meta AI, Llama2 is a series of foundation models trained on 2 trillion tokens from a diverse mix of publicly available sources~\cite{touvron_Llama_2023b}. Although Wikipedia is part of the pre-training data, the language distribution is extremely imbalanced, with approximately 90\% of the material in English. Compared to its predecessor, Llama2 has not only been trained on more data, but also features a doubled context length and grouped-query attention. We use Llama2-7B\footnote{\url{https://huggingface.co/meta-Llama/Llama-2-7b-hf}}. 
In order to base our observations on more than one model and to strengthen the findings of this study, we also experiment with Mistral, specifically Mistral-7B\footnote{\url{https://huggingface.co/mistralai/Mistral-7B-v0.3}}. This model is based on a similar architecture to Llama2 and integrates grouped-query and sliding window attention~\cite{jiang_mistral_2023}. Little information is published about its training data, except that it has been trained on several languages, with a focus on French, German, Spanish and Italian. Mistral outperforms Llama2 on several benchmarks.

\vspace{0.1cm}
\noindent \textbf{\bf Noise Augmented Token Dataset.}
For both representation-level and neuron-level experiments, we need the same input with varying levels of OCR noise. To this end, we build a fairly simple dataset composed of unique tokens, which are then replicated three times with different OCR noise levels. As the basis for correct, non-noisy text, we use the corrected version of French historical newspapers of the \textit{Common Corpus}\footnote{Released by PleIAs at \url{https://huggingface.co/datasets/PleIAs/Post-OCR-Correction}.}, from which we randomly select 2,447 (up to 400MB) documents. The text is tokenised, POS-tagged, and, by selecting only verbs and nouns, a set of 158,513 unique \texttt{correct} tokens is created. These tokens are then  altered by introducing character-level distortions typical of OCR engine errors, such as substituting similar-looking characters or omitting characters altogether. The error simulation is performed using the NLPAug library~\cite{ma_makcedward_2024} and is applied three times to each token with different parameters, resulting in pairs of $(token\_correct, token\_altered)$ with different OCR noise levels, as follows:
 
\begin{itemize}[topsep=0pt]
\setlength\itemsep{0.4em}
    \item \texttt{low}: insertion of one character alteration, e.g. $(editorial, editor1al)$, producing altered variants within 0.8--1.0 Levenshtein similarity;
    \item \texttt{average}: insertion of 2 to 5 character alterations, e.g. $(editorial, ed1tur1al)$ (0.6--0.8 Levenshtein similarity);
    \item \texttt{high}: insertion of 3 to 10 character alteration, e.g. $(editorial, eo1t0r1al)$ (0--0.6 Levenshtein similarity).
\end{itemize}

This generation method was favoured over using real OCR output as it allows for easier control of the noise level. The result is a multi-level noise augmented token dataset consisting of three sets of 158K pairs of $(correct, altered)$ tokens. These pairs form the basis of our experiments, where we passed them through the model components under investigation.

\subsection{Detecting Layer Regions Sensitive to OCR Noise}
\label{sec:sensitive-layers}

Our initial focus is on studying full vector representations to determine whether certain regions of the model are sensitive to OCR noise\footnote{The code relies on the Ecco \cite{alammar_ecco_2021} and on Captum (\url{https://captum.ai/}) libraries, and can be consulted at \url{https://github.com/EmanuelaBoros/ocr-sensitive-neurons}.}. 

\vspace{0.1cm}
\noindent \textbf{Network Structure.}
Llama2 and Mistral are based on the same architecture, which consists of an embedding layer, followed by 32 decoder blocks, and a final output layer~\cite{vaswani_attention_2017a}. Each decoder block includes a self-attention layer and a feed-forward neural network implemented as a multi-layer perceptron (MLP), both preceded by normalisation and followed by residual connections. The MLP layers consist of two fully connected layers with a non-linear activation in between: the first layer projects representations to a higher dimensional space (from 4,096 to 11,008), thereby enriching the feature sets by expanding the dimension of the hidden states \cite{alizadeh_llm_2024}, and the second projects the activation back to the original dimensions. Because they apply non-linear transformations to hidden states, MLP layers are essential to learning complex patterns and have been shown to act as memories encoding knowledge~\cite{dai_knowledge_2022,geva_transformer_2022a}.

\vspace{0.1cm}
\noindent \textbf{Experiment.} We focus on the MLP up-projection layers within each decoder block, examining their behaviour when fed with \texttt{correct} tokens compared with their counterparts with \texttt{high}, \texttt{average} and \texttt{low} levels of OCR noise. To measure the similarity or dissimilarity between two layers, we use the linear centered kernel alignment (CKA) similarity index~\cite{kornblith_similarity_2019}. CKA is specifically designed to compare internal representations of neural networks based on layer activations, both within a single network and across different networks. 
Invariant to orthogonal transformations and scaling, CKA is particularly helpful in neural network interpretability and analysis as it allows to understand how different input variables, network architectures, or even training datasets influence internal representations of models. For two representations, a high CKA value (close to 1) indicates that they are very similar, while a low CKA value (close to 0) reveals activation variability, indicating dissimilarity and suggesting that the layers or models process information differently.

\sloppy In this experiment, each token and its three altered versions are fed (forward pass) through each up-projection layer of the 32 decoders in our LLMs, and the CKA value is computed on the obtained activations (\textit{CKA(layer\_activations(token\_correct), layer\_activations(token\_altered))}). This process provides, for each up-projection layers, a measure of how similarly the layer responds to a correct token compared to the three altered versions. 

\begin{figure*}[t]
    \centering
    \includegraphics[width=1.\textwidth]{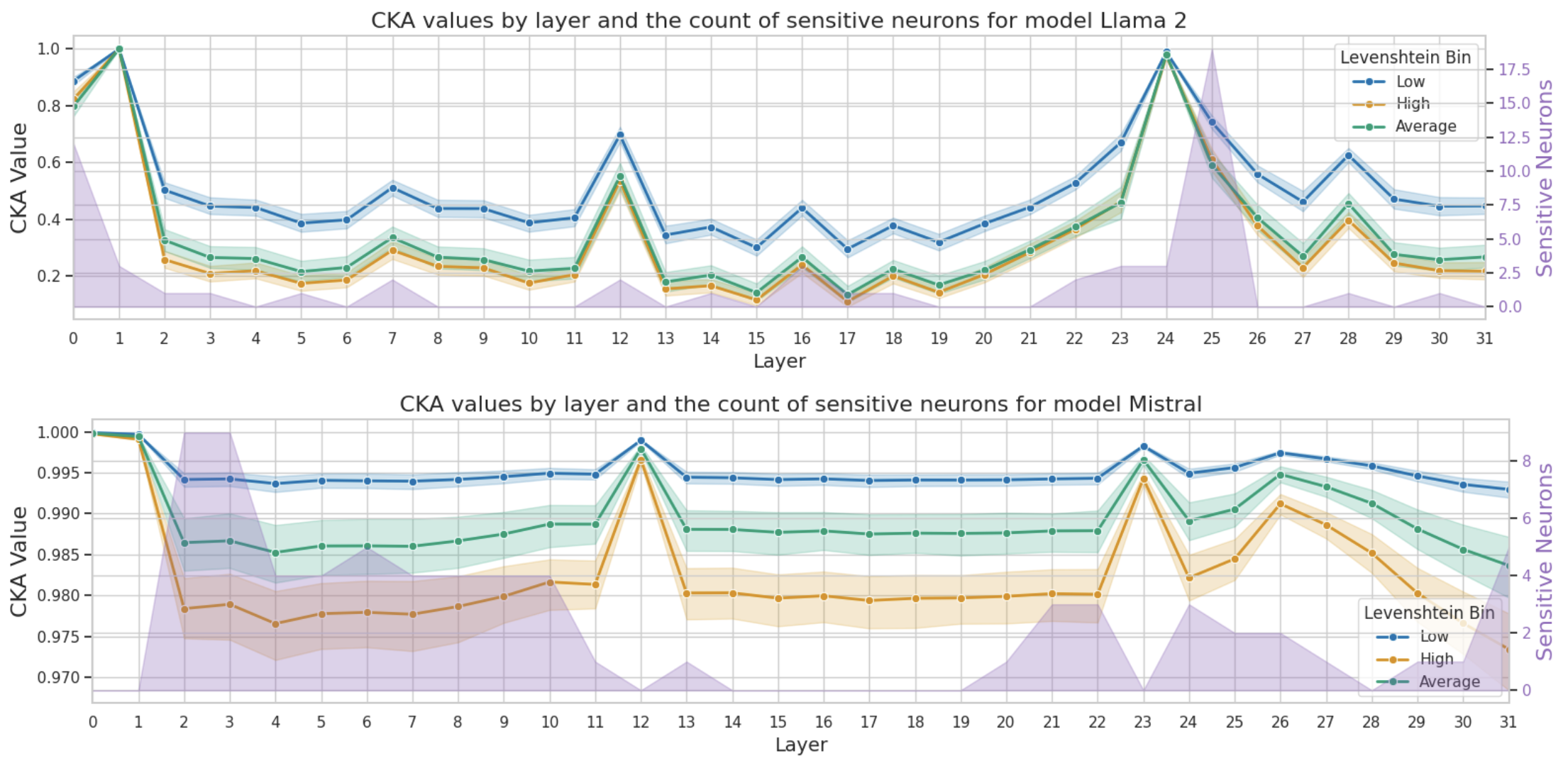}
    \caption{This figure shows 1) The CKA values (y-axis) between correct and \texttt{low|average|high} OCR noise-altered input tokens and 2) the number of OCR-sensitive neurons (secondary y-axis) for each of the 32 layers (x-axis) for Llama (top) and Mistral (bottom).}
    \label{fig:ocr_sensitive_neurons}
\end{figure*}

\vspace{0.1cm}
\noindent \textbf{Results.}
Figure \ref{fig:ocr_sensitive_neurons} shows the CKA values between correct and \texttt{low|average|high} OCR noise-altered input tokens for each of the 32 layers. Specifically, a given CKA value point at a given layer corresponds to the CKA between the activation of that layer in response to a correct token and the activation of the same layer in response to one of the altered versions of the same token. 
A first observation is that some layers are sensitive to OCR noise, and that this sensitivity varies depending on the layer's position in the network and on the noise level. For both models, the CKA values between \texttt{correct-altered} tokens are low for layers 2-11, 13-23 and 27-31, indicating that their activation values are far apart. In contrast, early layers (0--2), a middle layer (12), and a later layer (23) have CKA values close to 1, indicating similar activation and high consistency between layers. These peaks could indicate that these layers play a role in dealing with OCR. For all layers, we also note that the most severe noise level corresponds to the lowest CKA values.

A second observation is that both LLMs seem to react similarly, with CKA variations across layers and noise levels following the same dynamics. These commonalities suggest that these regions are consistently sensitive or not to OCR noise, regardless of the language of the LLM's pre-training data.  However, it should be noted that, despite similar trends, levels differ: Mistral has higher CKA values (between 0.97 and 1) across  layers compared to Llama2, suggesting a  greater robustness to OCR noise, possibly due to its knowledge of French. Llama2 shows a broader range of CKA values with more variability and lower minimum values (around 0.2), indicating a higher level of sensitivity to OCR noise.

Overall, this experiment shows that among the MLP layers of both Llama2 and Mistral, certain regions, particularly in the middle (2-11 and 13-23), exhibit significant variability in activations in response to correct and altered input, indicating higher sensitivity to OCR noise.

\subsection{Identifying OCR-sensitive Neurons}
\label{sec:sensitive-neurons}

We now focus on finding  neurons that react to OCR noise by using their activation differences.

\vspace{0.1cm}
\noindent \textbf{Neurons.}
In the layers of the MLP neural network, it is possible to study the activation of neurons by examining the transformations that occur through the layers, particularly before and after the application of the non-linear activation function. More specifically, the functioning of neurons can be described by considering how an MLP processes a normalised hidden state vector $x \in \mathbb{R}^{d_{\text{model}}}$, that is to say: $\text{MLP}(x) = W_{\text{out}} \sigma(W_{\text{in}} x + b_{\text{in}}) + b_{\text{out}}$ 
where \( W_{\text{out}} \) and \( W_{\text{in}} \) are matrices in \( \mathbb{R}^{d_{\text{model}} \times d_{\text{mlp}}} \) and \( \mathbb{R}^{d_{\text{mlp}} \times d_{\text{model}}} \), respectively, that are learned over time, and $b_{\text{in}}$ and $b_{\text{out}}$ are the learned bias terms. The function $\sigma$ is a point-wise non-linear activation function, corresponding to SwiGLU in the models analysed here~\cite{shazeer_glu_2020}. The activation of a specific neuron $j$ for various inputs $x$ can be examined using $\sigma(w_{\text{in}}^j x + b_{\text{in}}^j)$. Similarly, the activations of a neuron can be examined by looking at row $j$ of $W_{\text{in}}$ or the transpose of $W_{\text{out}}$. 

\vspace{0.1cm}
\noindent \textbf{Experiment.} 
Similarly to the previous experiment, we provide each MLP layer with pairs of \texttt{correct} and \texttt{altered} tokens. Instead of examining layer activations via CKA, we focus on differences in neuron activations to identify neurons that significantly and consistently ``deviate'' from others when exposed to OCR noise. First, a neuron \( n \) in an MLP layer is considered activated if its activation values $\sigma(x)$ exceed zero~\cite{nair_rectified_2010,tang_languagespecific_2024a}. Then, it is defined as OCR-sensitive if the difference between its activations in response to \texttt{correct} tokens and its activations in response to \texttt{altered} tokens is significantly higher than that of other neurons in the same layer when responding to the same inputs. The activation difference $\text{act\_diff}_n$ of neuron \( n \) is the absolute difference between the mean of its activations for \texttt{correct} and \texttt{altered} tokens: $\text{diff}_n = |\mu_{\text{correct\_token(s)}} - \mu_{\text{altered\_token(s)}}|$. 
This difference is significant if it is greater than the sum of the mean activation differences \( \mu_{\text{diff}} \) and the standard deviation of the activation differences $\sigma_{\text{diff}}$ for all other neurons in the layer: $\text{diff}_n > \mu_{\text{diff}} + \sigma_{\text{diff}}$. We compute the activation differences of neurons in each layer across all pairs and identify those with significant differences. Neurons showing significant activation differences in more than 90\% of the pairs are classified as OCR-sensitive neurons. This conservative threshold helps identify neurons that consistently react strongly to OCR noise across \texttt{altered} tokens.

\vspace{0.1cm}
\noindent \textbf{Results.} 
The number of OCR-sensitive neurons for each layer in both models is shown in Figure \ref{fig:ocr_sensitive_neurons} ( purple shaded area), together with the CKA values. We observe that the theoretical inverse relationship between a low CKA value and a higher number of neurons classified as OCR-sensitive holds true, albeit with some exceptions. If this negative relationship is clear for the early layers (2-10) of Mistral, for layers 12 and some later layers (24-26) of both models, the observation points for Llama 2 are less consistent. This shows that while these approaches appear to be effective in assessing the presence of OCR-sensitive layers and neurons, it remains difficult to obtain unambiguous observations on which to base clear-cut conclusions for large models. Nevertheless, both the CKA values and the number of OCR-sensitive neurons show that there are groups of layers and neurons that are consistently sensitive to OCR noise, suggesting that it may be worth trying to neutralise neurons to observe the effect on models' performance on NER on noisy documents. 
\vspace{-0.4cm}


\section{Neuron Ablation for NER on Historical Documents}
\label{sec:downstream-task}

We investigate the influence of OCR-sensitive neurons on LLM's performance in NER on historical documents through a systematic neuron ablation study.

\begin{figure*}[t]
    \centering
    \includegraphics[width=.9\textwidth]{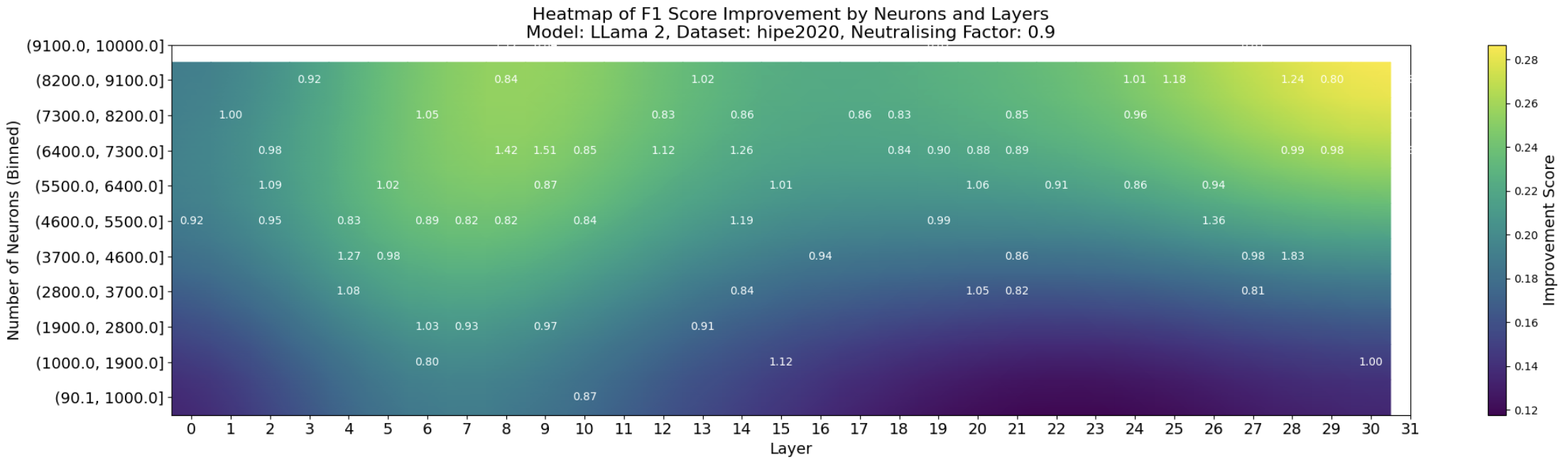}
    \includegraphics[width=.9\textwidth]{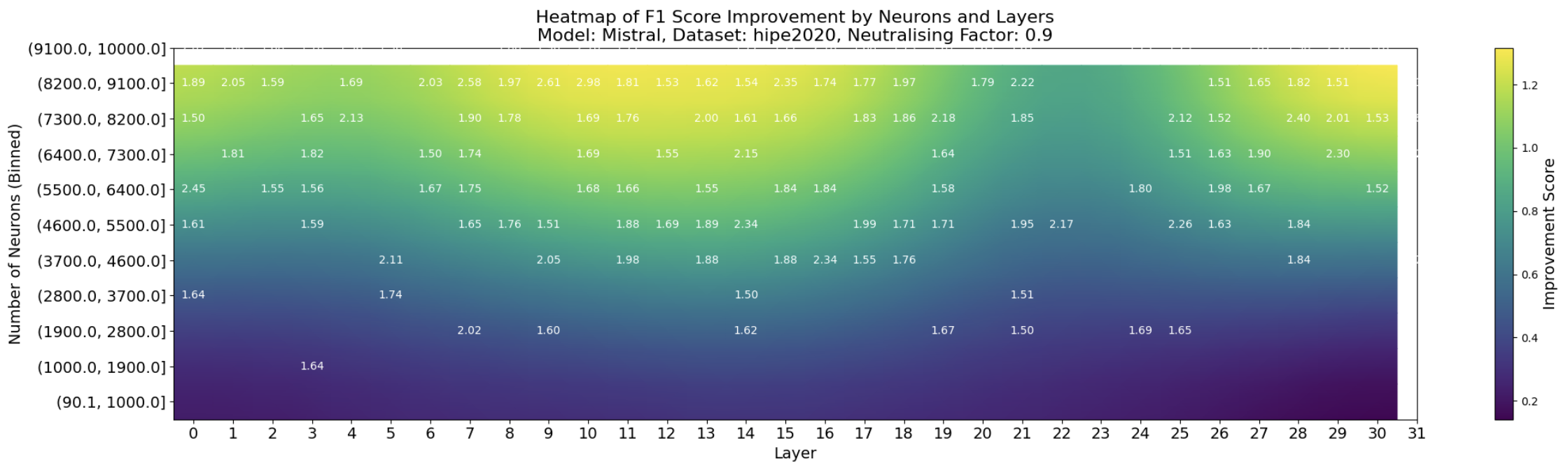}
    \caption{F1 score improvements with neuron activation modifications by neuron bins (y-axis) and layers (x-axis) of Llama2 and Mistral on \textbf{\texttt{hipe2020}} with a neutralising factor of 0.9. Warmer colour indicates higher F1 score improvement.}
    \label{fig:hipe_improvement_heatmap}
\end{figure*}

\begin{figure*}[h]
    \centering
    \includegraphics[width=.9\textwidth]{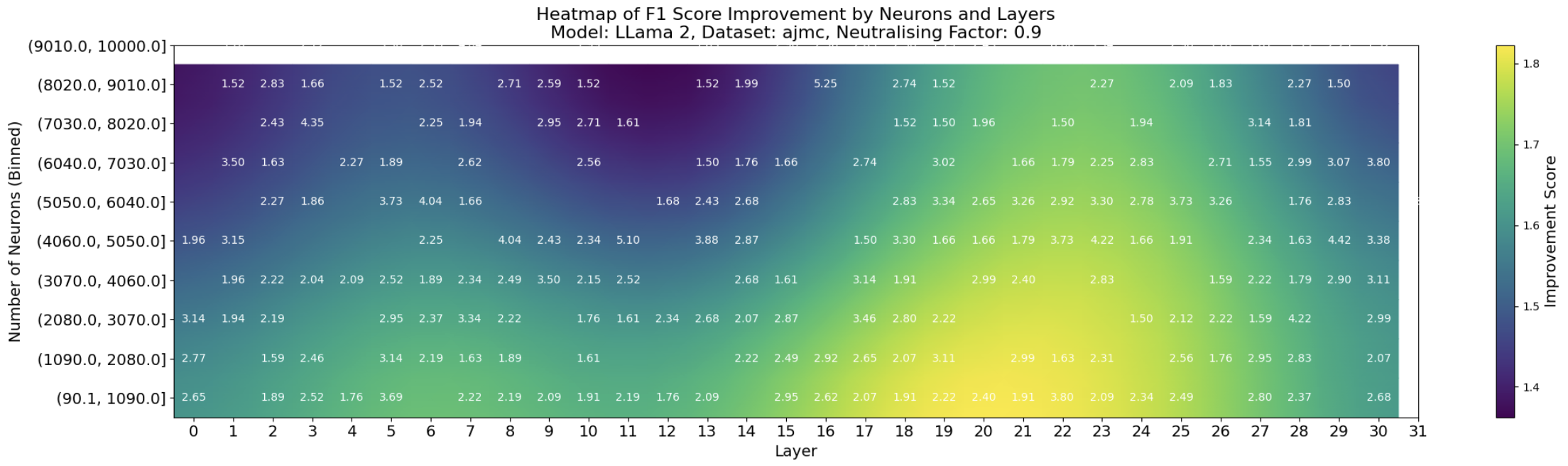}
    \includegraphics[width=.9\textwidth]{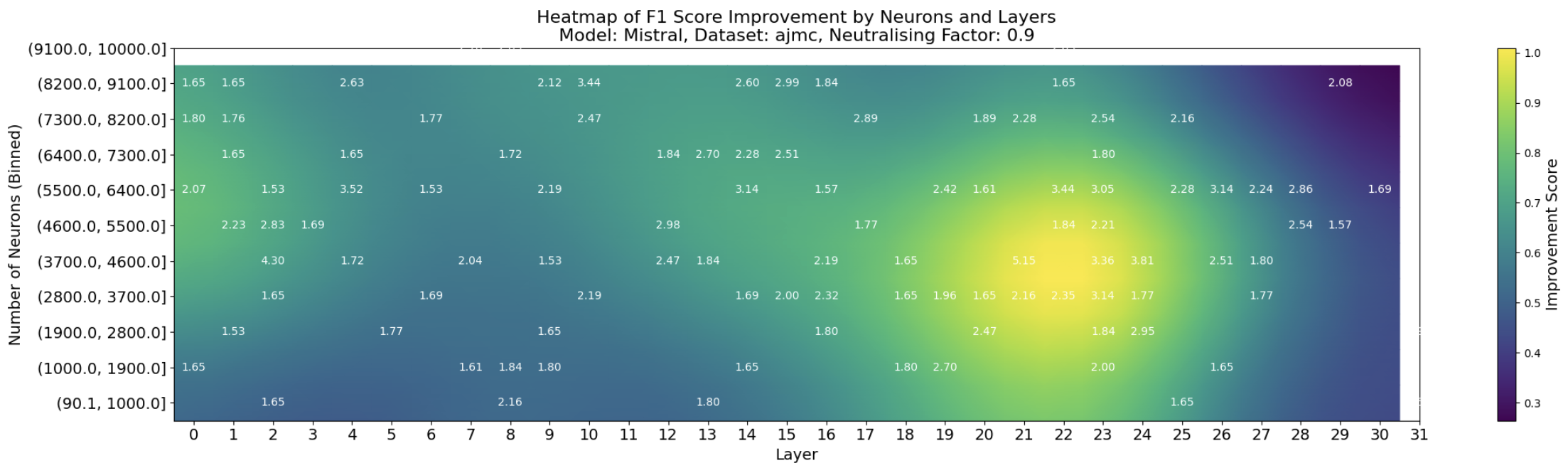}
    \caption{F1-score improvements of Llama2 (top) and Mistral (bottom) on \textbf{\texttt{ajmc}}.}

    \label{fig:ajmc_improvement_heatmap}
\end{figure*}


\vspace{0.1cm}
\noindent \textbf{\bf Data.}
We use the French parts of the HIPE-2020 (\texttt{hipe2020}) and of the Ajax Multi-Commentary (\texttt{ajmc}) NER benchmarks~\cite{ehrmann_extended_2022b,romanello_named_2024}. \texttt{Hipe2020} consists of historical newspapers from 19C-20C, while \texttt{ajmc} includes 19C scholarly commentaries on Sophocles' \textit{Ajax}. Both datasets contain OCR noise.

\vspace{0.1cm}
\noindent \textbf{\bf Experiments.} 
Both models are fine-tuned using the law-rank adaptation strategy \cite{hu_lora_2021}\footnote{Please refer to Appendix \ref{appendix:finetuning} for more details.}. To establish a baseline, models are fine-tuned on the training sets without neuron ablation. Each fine-tuning process is repeated five times and the results on test sets are averaged to account for possible variances between runs. Baseline F1-scores are: for \texttt{hipe2020}, $50.72\pm0.07$ (Llama2)  and $52.88\pm0.07$ (Mistral) and, for \texttt{ajmc}, $33.63\pm0.05$ (Llama2) and $46.29\pm0.06$ (Mistral).
Next, we neutralise neurons of the MLP up-projection layers by changing their activation values during the forward pass. For each neuron \( n \), activations are adjusted by introducing a mask \( M \) containing a neutralising factor \( \alpha \), with $M[n] = \alpha$ if  \( n \) is a neuron to  neutralise and $M[n] = 1$ otherwise.
The adjusted hidden states \( h' \) are then given by \( h' = h \odot M \), where \( h \) represents the original hidden states and \( \odot \) denotes element-wise multiplication.

Although it is not possible to know in advance which exact neurons play a role in dealing with OCR and should be neutralised, the objective is to measure their importance and/or effect based on the magnitude changes in F1-scores. We neutralise between 100 to 11,000 neurons per layer in steps of 100 during the forward pass on the test data, using 0.1, 0.5, and 0.9 as neutralising factor values (reducing activations by 90\%, 50\%, and 10\%, respectively). Neuron neutralisation effect is evaluated via the change in F1-score, computed as $\frac{\text{Neutralised F1} - \text{Baseline F1}}{\text{Baseline F1}} \times 100$.

\vspace{0.1cm}
\noindent \textbf{\bf Results.} Improvements of F1-scores are shown in Figures \ref{fig:hipe_improvement_heatmap} and \ref{fig:ajmc_improvement_heatmap} for \texttt{hipe2020} and \texttt{ajmc} datasets with a neutralising factor of 0.9\footnote{We found that a neutralising factor of 0.9 leads to higher F1-score improvements for both datasets; please refer to  Appendix \ref{appendix:improvement_scores_per_dataset}.}. 
For \texttt{ajmc}, we observe that neutralising a small number of neurons (< 3,000 neuron bins) in layers 18-24 for Llama2 and layers 20-24 for Mistral leads to significant improvements of about 3\% for Llama2 and 5\% for Mistral, suggesting that layers 21-23 are of similar importance for both LLMs. A different set of layers comes into play for \texttt{hipe2020}, with higher improvements achieved by neutralising a larger number of neurons (> 6,400) in the middle and late layers of the models. While this differs from the results for \texttt{ajmc}, it is consistent with the distribution of CKA values. We can conclude that optimising layers 20-24 for Llama 2 and 29-31 for Mistral by enhancing their representations could enhance their robustness to OCR errors and yield performance gains. 
\vspace{-0.5cm}

\section{Conclusions}
This study provides a preliminary assessment of OCR-sensitive layers and neurons in Transformer models, highlighting their impact on model performance in the presence of OCR. We identified critical layer regions within Llama 2 and Mistral that exhibit heightened sensitivity to OCR noise. Optimising these layers (0-2, 11-13, 23-28) could enhance NER performance, particularly in the final layers. However, the distinct improvements observed across different datasets (\texttt{hipe2020} and \texttt{ajmc}) underscore the need for noise level-specific adaptations. Future work will focus on  specific types of OCR errors and their distributions in the training data. Additionally, expanding the analysis to other models and datasets can provide a broader perspective on OCR sensitivity and performance improvements in various contexts.

\section*{Limitations}

Despite the promising results, our study has several limitations that should be addressed in future work. First, the process of identifying OCR-sensitive neurons is highly dependent on the specific datasets and noise levels used in the experiments. While we used a diverse set of OCR noise levels, we focused only on two datasets in French, thus the findings may not generalise across different OCR systems or text corpora. Not only, but we considered only nouns and verbs for this preliminary experiments, while probably missing activations on other parts of speech. Thus, we leave this for our future work, where we will explore more datasets and OCR noise conditions to validate the generalisability of the identified sensitive neurons. Second, the method of neuron neutralisation, which involves scaling down neuron activations, may oversimplify the complex dynamics of neuron interactions within the model. While this approach provides insights into the role of specific neurons, it does not account for potential compensatory mechanisms that might occur when certain neurons are deactivated. More sophisticated techniques, such as neuron editing or neuron replacement strategies, could provide a deeper understanding of the role of these neurons. Finally, while we showed improvements in model performance by neutralising OCR-sensitive neurons, the practical implementation of these techniques in real-world systems remains challenging. We would still need to address the computational overhead and complexity of dynamically identifying and neutralising these neurons.

\subsubsection{\ackname} 
 Authors gratefully acknowledge the financial support of the Swiss National Science Foundation (SNSF) for the research projects `\textit{Impresso} - Media Monitoring of the Past II. Beyond Borders: Connecting Newspaper and Radio`.



%
\bibliographystyle{splncs04}
\bibliography{2024-ocr-sensitive-neurons}

\begin{thebibliography}{10}
\providecommand{\url}[1]{\texttt{#1}}
\providecommand{\urlprefix}{URL }
\providecommand{\doi}[1]{https://doi.org/#1}

\bibitem{adi_finegrained_2017}
Adi, Y., Kermany, E., Belinkov, Y., Lavi, O., Goldberg, Y.: Fine-grained
  {{Analysis}} of {{Sentence Embeddings Using Auxiliary Prediction Tasks}} (Feb
  2017), \url{http://arxiv.org/abs/1608.04207}

\bibitem{alammar_ecco_2021}
Alammar, J.: Ecco: {{An Open Source Library}} for the {{Explainability}} of
  {{Transformer Language Models}}. In: Ji, H., Park, J.C., Xia, R. (eds.)
  Proceedings of the 59th {{Annual Meeting}} of the {{Association}} for
  {{Computational Linguistics}} and the 11th {{International Joint Conference}}
  on {{Natural Language Processing}}: {{System Demonstrations}}. pp. 249--257.
  Association for Computational Linguistics, Online (Aug 2021),
  \url{https://aclanthology.org/2021.acl-demo.30}

\bibitem{alizadeh_llm_2024}
Alizadeh, K., Mirzadeh, I., Belenko, D., Khatamifard, K., Cho, M., Del~Mundo,
  C.C., Rastegari, M., Farajtabar, M.: {{LLM}} in a flash: {{Efficient Large
  Language Model Inference}} with {{Limited Memory}} (Jan 2024),
  \url{http://arxiv.org/abs/2312.11514}

\bibitem{bau_identifying_2018}
Bau, A., Belinkov, Y., Sajjad, H., Durrani, N., Dalvi, F., Glass, J.:
  Identifying and {{Controlling Important Neurons}} in {{Neural Machine
  Translation}} (Nov 2018), \url{http://arxiv.org/abs/1811.01157}

\bibitem{belinkov_probing_2022b}
Belinkov, Y.: Probing {{Classifiers}}: {{Promises}}, {{Shortcomings}}, and
  {{Advances}}. Computational Linguistics  \textbf{48}(1),  207--219 (Apr
  2022), \url{https://doi.org/10.1162/coli_a_00422}

\bibitem{belinkov_linguistic_2020}
Belinkov, Y., Durrani, N., Dalvi, F., Sajjad, H., Glass, J.: On the
  {{Linguistic Representational Power}} of {{Neural Machine Translation
  Models}}. Computational Linguistics  \textbf{46}(1),  1--52 (Mar 2020),
  \url{https://doi.org/10.1162/coli_a_00367}

\bibitem{belinkov_analysis_2019}
Belinkov, Y., Glass, J.: Analysis {{Methods}} in {{Neural Language
  Processing}}: {{A Survey}}. Transactions of the Association for Computational
  Linguistics  \textbf{7},  49--72 (Apr 2019),
  \url{https://doi.org/10.1162/tacl_a_00254}

\bibitem{boros_postcorrection_2024}
Boros, E., Ehrmann, M., Romanello, M., {Najem-Meyer}, S., Kaplan, F.:
  Post-{{Correction}} of {{Historical Text Transcripts}} with {{Large Language
  Models}}: {{An Exploratory Study}}. In: Bizzoni, Y., {Degaetano-Ortlieb}, S.,
  Kazantseva, A., Szpakowicz, S. (eds.) Proceedings of the 8th {{Joint SIGHUM
  Workshop}} on {{Computational Linguistics}} for {{Cultural Heritage}},
  {{Social Sciences}}, {{Humanities}} and {{Literature}} ({{LaTeCH-CLfL}}
  2024). pp. 133--159. Association for Computational Linguistics, St. Julians,
  Malta (Mar 2024), \url{https://aclanthology.org/2024.latechclfl-1.14}

\bibitem{boros_alleviating_2020}
Boros, E., Hamdi, A., Linhares~Pontes, E., {Cabrera-Diego}, L.A., Moreno, J.G.,
  Sidere, N., Doucet, A.: Alleviating {{Digitization Errors}} in {{Named Entity
  Recognition}} for {{Historical Documents}}. In: Fern{\'a}ndez, R., Linzen, T.
  (eds.) Proceedings of the 24th {{Conference}} on {{Computational Natural
  Language Learning}}. pp. 431--441. Association for Computational Linguistics,
  Online (Nov 2020), \url{https://aclanthology.org/2020.conll-1.35}

\bibitem{candela_reusing_2022}
Candela, G., S{\'a}ez, M.D., Escobar~Esteban, {\relax Mp}., {Marco-Such}, M.:
  Reusing digital collections from {{GLAM}} institutions. Journal of
  Information Science  \textbf{48}(2),  251--267 (Apr 2022),
  \url{https://doi.org/10.1177/0165551520950246}

\bibitem{conneau_what_2018a}
Conneau, A., Kruszewski, G., Lample, G., Barrault, L., Baroni, M.: What you can
  cram into a single \$\&!\#* vector: {{Probing}} sentence embeddings for
  linguistic properties. In: Gurevych, I., Miyao, Y. (eds.) Proceedings of the
  56th {{Annual Meeting}} of the {{Association}} for {{Computational
  Linguistics}} ({{Volume}} 1: {{Long Papers}}). pp. 2126--2136. Association
  for Computational Linguistics, Melbourne, Australia (Jul 2018),
  \url{https://aclanthology.org/P18-1198}

\bibitem{dai_knowledge_2022}
Dai, D., Dong, L., Hao, Y., Sui, Z., Chang, B., Wei, F.: Knowledge {{Neurons}}
  in {{Pretrained Transformers}}. In: Muresan, S., Nakov, P., Villavicencio, A.
  (eds.) Proceedings of the 60th {{Annual Meeting}} of the {{Association}} for
  {{Computational Linguistics}} ({{Volume}} 1: {{Long Papers}}). pp.
  8493--8502. Association for Computational Linguistics, Dublin, Ireland (May
  2022), \url{https://aclanthology.org/2022.acl-long.581}

\bibitem{dalvi_what_2019}
Dalvi, F., Durrani, N., Sajjad, H., Belinkov, Y., Bau, A., Glass, J.: What {{Is
  One Grain}} of {{Sand}} in the {{Desert}}? {{Analyzing Individual Neurons}}
  in {{Deep NLP Models}}. In: Proceedings of the {{AAAI Conference}} on
  {{Artificial Intelligence}}. vol.~33, pp. 6309--6317 (Jul 2019),
  \url{https://ojs.aaai.org/index.php/AAAI/article/view/4592}

\bibitem{dalvi_analyzing_2020}
Dalvi, F., Sajjad, H., Durrani, N., Belinkov, Y.: Analyzing {{Redundancy}} in
  {{Pretrained Transformer Models}}. In: Webber, B., Cohn, T., He, Y., Liu, Y.
  (eds.) Proceedings of the 2020 {{Conference}} on {{Empirical Methods}} in
  {{Natural Language Processing}} ({{EMNLP}}). pp. 4908--4926. Association for
  Computational Linguistics, Online (Nov 2020),
  \url{https://aclanthology.org/2020.emnlp-main.398}

\bibitem{devlin_bert_2019}
Devlin, J., Chang, M.W., Lee, K., Toutanova, K.: {{BERT}}: {{Pre-training}} of
  deep bidirectional transformers for language understanding. In: Proceedings
  of the 2019 Conference of the North {{American}} Chapter of the Association
  for Computational Linguistics: {{Human}} Language Technologies, Volume 1
  (Long and Short Papers). pp. 4171--4186. Association for Computational
  Linguistics, Minneapolis, Minnesota (Jun 2019). \doi{10.18653/v1/N19-1423}

\bibitem{doucet_newseye_2020}
Doucet, A., Gasteiner, M., {Granroth-Wilding}, M., Kaiser, M., Kaukonen, M.,
  Labahn, R., Moreux, J.P., Muehlberger, G., Pfanzelter, E., Th{\'e}renty,
  M.{\`E}., Toivonen, H., Tolonen, M.: {{NewsEye}}: {{A}} digital investigator
  for historical newspapers. In: 15th {{Annual International Conference}} of
  the {{Alliance}} of {{Digital Humanities Organizations}}, {{DH}} 2020 (Jul
  2020), \url{https://hal.science/hal-03029072}

\bibitem{durrani_analyzing_2020}
Durrani, N., Sajjad, H., Dalvi, F., Belinkov, Y.: Analyzing {{Individual
  Neurons}} in {{Pre-trained Language Models}}. In: Webber, B., Cohn, T., He,
  Y., Liu, Y. (eds.) Proceedings of the 2020 {{Conference}} on {{Empirical
  Methods}} in {{Natural Language Processing}} ({{EMNLP}}). pp. 4865--4880.
  Association for Computational Linguistics, Online (Nov 2020),
  \url{https://aclanthology.org/2020.emnlp-main.395}

\bibitem{ehrmann_named_2023a}
Ehrmann, M., Hamdi, A., Pontes, E.L., Romanello, M., Doucet, A.: Named {{Entity
  Recognition}} and {{Classification}} in {{Historical Documents}}: {{A
  Survey}}. ACM Comput. Surv.  \textbf{56}(2),  27:1--27:47 (Sep 2023),
  \url{https://doi.org/10.1145/3604931}

\bibitem{ehrmann_language_2020}
Ehrmann, M., Romanello, M., Clematide, S., Str{\"o}bel, P.B., Barman, R.:
  Language {{Resources}} for {{Historical Newspapers}}: The {{Impresso
  Collection}}. In: Calzolari, N., B{\'e}chet, F., Blache, P., Choukri, K.,
  Cieri, C., Declerck, T., Goggi, S., Isahara, H., Maegaard, B., Mariani, J.,
  Mazo, H., Moreno, A., Odijk, J., Piperidis, S. (eds.) Proceedings of the
  {{Twelfth Language Resources}} and {{Evaluation Conference}}. pp. 958--968.
  European Language Resources Association, Marseille, France (May 2020),
  \url{https://aclanthology.org/2020.lrec-1.121}

\bibitem{ehrmann_extended_2020b}
Ehrmann, M., Romanello, M., Fl{\"u}ckiger, A., Clematide, S.: Extended
  {{Overview}} of {{CLEF HIPE}} 2020: {{Named Entity Processing}} on
  {{Historical Newspapers}}. In: Cappellato, L., Eickhoff, C., Ferro, N.,
  N{\'e}v{\'e}ol, A. (eds.) Working {{Notes}} of {{CLEF}} 2020 - {{Conference}}
  and {{Labs}} of the {{Evaluation Forum}}. vol.~2696, p.~38. CEUR-WS,
  Thessaloniki, Greece (2020), \url{https://infoscience.epfl.ch/record/281054}

\bibitem{ehrmann_extended_2022b}
Ehrmann, M., Romanello, M., {Najem-Meyer}, S., Doucet, A., Clematide, S.:
  Extended {{Overview}} of {{HIPE-2022}}: {{Named Entity Recognition}} and
  {{Linking}} in {{Multilingual Historical Documents}}. In: Proceedings of the
  {{Working Notes}} of {{CLEF}} 2022 - {{Conference}} and {{Labs}} of the
  {{Evaluation Forum}}. CEUR-WS (2022). \doi{10.5281/zenodo.6979577}

\bibitem{ehrmann_overview_2022}
Ehrmann, M., Romanello, M., {Najem-Meyer}, S., Doucet, A., Clematide, S.:
  Overview of~{{HIPE-2022}}: {{Named Entity Recognition}} and~{{Linking}}
  in~{{Multilingual Historical Documents}}. In: {Barr{\'o}n-Cede{\~n}o}, A.,
  Da~San~Martino, G., Degli~Esposti, M., Sebastiani, F., Macdonald, C., Pasi,
  G., Hanbury, A., Potthast, M., Faggioli, G., Ferro, N. (eds.) Experimental
  {{IR Meets Multilinguality}}, {{Multimodality}}, and {{Interaction}}. pp.
  423--446. Lecture {{Notes}} in {{Computer Science}}, Springer International
  Publishing, Cham (2022). \doi{10.1007/978-3-031-13643-6_26}

\bibitem{erhan_visualizing_2009}
Erhan, D., Bengio, Y., Courville, A.C., Vincent, P.: Visualizing {{Higher-Layer
  Features}} of a {{Deep Network}} (2009),
  \url{https://api.semanticscholar.org/CorpusID:15127402}

\bibitem{geva_transformer_2022a}
Geva, M., Caciularu, A., Wang, K., Goldberg, Y.: Transformer {{Feed-Forward
  Layers Build Predictions}} by {{Promoting Concepts}} in the {{Vocabulary
  Space}}. In: Goldberg, Y., Kozareva, Z., Zhang, Y. (eds.) Proceedings of the
  2022 {{Conference}} on {{Empirical Methods}} in {{Natural Language
  Processing}}. pp. 30--45. Association for Computational Linguistics, Abu
  Dhabi, United Arab Emirates (Dec 2022),
  \url{https://aclanthology.org/2022.emnlp-main.3}

\bibitem{gurnee_language_2024a}
Gurnee, W., Tegmark, M.: Language {{Models Represent Space}} and {{Time}} (Mar
  2024), \url{http://arxiv.org/abs/2310.02207}

\bibitem{hamdi_assessing_2020a}
Hamdi, A., {Jean-Caurant}, A., Sid{\`e}re, N., Coustaty, M., Doucet, A.:
  Assessing and {{Minimizing}} the {{Impact}} of {{OCR Quality}} on {{Named
  Entity Recognition}}. In: Hall, M., Mer{\v c}un, T., Risse, T., Duchateau, F.
  (eds.) Digital {{Libraries}} for {{Open Knowledge}}. pp. 87--101. Lecture
  {{Notes}} in {{Computer Science}}, Springer International Publishing, Cham
  (2020). \doi{10.1007/978-3-030-54956-5_7}

\bibitem{hu_lora_2021}
Hu, E.J., Shen, Y., Wallis, P., {Allen-Zhu}, Z., Li, Y., Wang, S., Wang, L.,
  Chen, W.: {{LoRA}}: {{Low-Rank Adaptation}} of {{Large Language Models}} (Oct
  2021), \url{http://arxiv.org/abs/2106.09685}

\bibitem{hupkes_visualisation_2018}
Hupkes, D., Veldhoen, S., Zuidema, W.: Visualisation and '{{Diagnostic
  Classifiers}}' {{Reveal How Recurrent}} and {{Recursive Neural Networks
  Process Hierarchical Structure}}. Journal of Artificial Intelligence Research
   \textbf{61},  907--926 (Apr 2018),
  \url{https://www.jair.org/index.php/jair/article/view/11196}

\bibitem{huynh_when_2020a}
Huynh, V.N., Hamdi, A., Doucet, A.: When to {{Use OCR Post-correction}} for
  {{Named Entity Recognition}}? In: Ishita, E., Pang, N.L.S., Zhou, L. (eds.)
  Digital {{Libraries}} at {{Times}} of {{Massive Societal Transition}}. pp.
  33--42. Springer International Publishing, Cham (2020).
  \doi{10.1007/978-3-030-64452-9_3}

\bibitem{jiang_mistral_2023}
Jiang, A.Q., Sablayrolles, A., Mensch, A., Bamford, C., Chaplot, D.S., de~las
  Casas, D., Bressand, F., Lengyel, G., Lample, G., Saulnier, L., Lavaud, L.R.,
  Lachaux, M.A., Stock, P., Scao, T.L., Lavril, T., Wang, T., Lacroix, T.,
  Sayed, W.E.: Mistral {{7B}} (Oct 2023), \url{http://arxiv.org/abs/2310.06825}

\bibitem{karpathy_visualizing_2015}
Karpathy, A., Johnson, J., {Fei-Fei}, L.: Visualizing and {{Understanding
  Recurrent Networks}} (Nov 2015), \url{http://arxiv.org/abs/1506.02078}

\bibitem{kornblith_similarity_2019}
Kornblith, S., Norouzi, M., Lee, H., Hinton, G.: Similarity of {{Neural Network
  Representations Revisited}}. In: Proceedings of the 36th {{International
  Conference}} on {{Machine Learning}}. pp. 3519--3529. PMLR (May 2019),
  \url{https://proceedings.mlr.press/v97/kornblith19a.html}

\bibitem{li_label_2023c}
Li, Z., Li, X., Liu, Y., Xie, H., Li, J., Wang, F.l., Li, Q., Zhong, X.: Label
  {{Supervised LLaMA Finetuning}} (Oct 2023),
  \url{http://arxiv.org/abs/2310.01208}

\bibitem{linharespontes_impact_2019a}
Linhares~Pontes, E., Hamdi, A., Sidere, N., Doucet, A.: Impact of {{OCR
  Quality}} on {{Named Entity Linking}}. In: Jatowt, A., Maeda, A., Syn, S.Y.
  (eds.) Digital {{Libraries}} at the {{Crossroads}} of {{Digital Information}}
  for the {{Future}}. pp. 102--115. Lecture {{Notes}} in {{Computer Science}},
  Springer International Publishing, Cham (2019).
  \doi{10.1007/978-3-030-34058-2_11}

\bibitem{ma_makcedward_2024}
Ma, E.: Makcedward/nlpaug (Jul 2024),
  \url{https://github.com/makcedward/nlpaug}

\bibitem{manjavacas_adapting_2022}
Manjavacas, E., Fonteyn, L.: Adapting vs. {{Pre-training Language Models}} for
  {{Historical Languages}}. Journal of Data Mining and Digital Humanities
  \textbf{NLP4DH} (Jun 2022), \url{https://inria.hal.science/hal-03592137}

\bibitem{mcgillivray_challenges_2020}
McGillivray, B., Alex, B., Ames, S., Armstrong, G., Beavan, D., Ciula, A.,
  Colavizza, G., Cummings, J., Roure, D.D., Farquhar, A., Hengchen, S., Lang,
  A., Loxley, J., Goudarouli, E., Nanni, F., Nini, A., Nyhan, J., Osborne, N.,
  Poibeau, T., Ridge, M., Ranade, S., Smithies, J., Terras, M., Vlachidis, A.,
  Willcox, P.: The challenges and prospects of the intersection of humanities
  and data science: {{A White Paper}} from {{The Alan Turing Institute}}. Tech.
  rep., Alan Turing Institute (Aug 2020),
  \url{https://figshare.com/articles/online_resource/The_challenges_and_prospects_of_the_intersection_of_humanities_and_data_science_A_White_Paper_from_The_Alan_Turing_Institute/12732164}

\bibitem{moradi_evaluating_2021}
Moradi, M., Samwald, M.: Evaluating the {{Robustness}} of {{Neural Language
  Models}} to {{Input Perturbations}}. In: Moens, M.F., Huang, X., Specia, L.,
  Yih, S.W.t. (eds.) Proceedings of the 2021 {{Conference}} on {{Empirical
  Methods}} in {{Natural Language Processing}}. pp. 1558--1570. Association for
  Computational Linguistics, Online and Punta Cana, Dominican Republic (Nov
  2021), \url{https://aclanthology.org/2021.emnlp-main.117}

\bibitem{na_discovery_2019}
Na, S., Choe, Y.J., Lee, D.H., Kim, G.: Discovery of {{Natural Language
  Concepts}} in {{Individual Units}} of {{CNNs}} (Feb 2019),
  \url{http://arxiv.org/abs/1902.07249}

\bibitem{nair_rectified_2010}
Nair, V., Hinton, G.E.: Rectified linear units improve restricted boltzmann
  machines. In: Proceedings of the 27th International Conference on Machine
  Learning ({{ICML-10}}). pp. 807--814 (2010)

\bibitem{neudecker_ocrd_2019}
Neudecker, C., Baierer, K., Federbusch, M., Boenig, M., W{\"u}rzner, K.M.,
  Hartmann, V., Herrmann, E.: {{OCR-D}}: {{An}} end-to-end open source {{OCR}}
  framework for historical printed documents. In: Proceedings of the 3rd
  {{International Conference}} on {{Digital Access}} to {{Textual Cultural
  Heritage}} - {{DATeCH2019}}. pp. 53--58. ACM Press, Brussels, Belgium (2019),
  \url{http://dl.acm.org/citation.cfm?doid=3322905.3322917}

\bibitem{padilla_responsible_2020}
Padilla, T.: Responsible {{Operations}}: {{Data Science}}, {{Machine
  Learning}}, and {{AI}} in {{Libraries}}. Tech. rep., OCLC (May 2020),
  \url{https://www.oclc.org/content/research/publications/2019/oclcresearch-responsible-operations-data-science-machine-learning-ai.html}

\bibitem{rehm_qurator_2020}
Rehm, G., Bourgonje, P., Hegele, S., Kintzel, F., Schneider, J.M., Ostendorff,
  M., Zaczynska, K., Berger, A., Grill, S., R{\"a}uchle, S., Rauenbusch, J.,
  Rutenburg, L., Schmidt, A., Wild, M., Hoffmann, H., Fink, J., Schulz, S.,
  Seva, J., Quantz, J., B{\"o}ttger, J., Matthey, J., Fricke, R., Thomsen, J.,
  Paschke, A., Qundus, J.A., Hoppe, T., Karam, N., Weichhardt, F., Fillies, C.,
  Neudecker, C., Gerber, M., Labusch, K., Rezanezhad, V., Schaefer, R.,
  Zellh{\"o}fer, D., Siewert, D., Bunk, P., Pintscher, L., Aleynikova, E.,
  Heine, F.: {{QURATOR}}: {{Innovative Technologies}} for {{Content}} and
  {{Data Curation}} (Apr 2020), \url{http://arxiv.org/abs/2004.12195}

\bibitem{romanello_named_2024}
Romanello, M., {Najem-Meyer}, S.: A {{Named Entity-Annotated Corpus}} of 19th
  {{Century Classical Commentaries}}. Journal of Open Humanities Data
  \textbf{10}(1) (Jan 2024),
  \url{https://openhumanitiesdata.metajnl.com/articles/10.5334/johd.150}

\bibitem{sajjad_neuronlevel_2022}
Sajjad, H., Durrani, N., Dalvi, F.: Neuron-level {{Interpretation}} of {{Deep
  NLP Models}}: {{A Survey}}. Transactions of the Association for Computational
  Linguistics  \textbf{10},  1285--1303 (Nov 2022),
  \url{https://doi.org/10.1162/tacl_a_00519}

\bibitem{sajjad_analyzing_2022a}
Sajjad, H., Durrani, N., Dalvi, F., Alam, F., Khan, A., Xu, J.: Analyzing
  {{Encoded Concepts}} in {{Transformer Language Models}}. In: Carpuat, M., {de
  Marneffe}, M.C., Meza~Ruiz, I.V. (eds.) Proceedings of the 2022
  {{Conference}} of the {{North American Chapter}} of the {{Association}} for
  {{Computational Linguistics}}: {{Human Language Technologies}}. pp.
  3082--3101. Association for Computational Linguistics, Seattle, United States
  (Jul 2022), \url{https://aclanthology.org/2022.naacl-main.225}

\bibitem{schweter_hmbert_2022a}
Schweter, S., M{\"a}rz, L., Schmid, K., {\c C}ano, E.: {{hmBERT}}: {{Historical
  Multilingual Language Models}} for {{Named Entity Recognition}}. In:
  Faggioli, G., Ferro, N., Hanbury, A., Potthast, M. (eds.) Proceedings of the
  {{Working Notes}} of {{CLEF}} 2022 - {{Conference}} and {{Labs}} of the
  {{Evaluation Forum}}. {{CEUR Workshop Proceedings}}, vol.~3180, pp.
  1109--1129. CEUR, Bologna, Italy (Sep 2022),
  \url{http://ceur-ws.org/Vol-3180/#paper-87}

\bibitem{shazeer_glu_2020}
Shazeer, N.: {{GLU Variants Improve Transformer}} (Feb 2020),
  \url{http://arxiv.org/abs/2002.05202}

\bibitem{smith_computational_2015}
Smith, D.A., Cordell, R., Mullen, A.: Computational {{Methods}} for
  {{Uncovering Reprinted Texts}} in {{Antebellum Newspapers}}. American
  Literary History  \textbf{27}(3),  E1--E15 (Sep 2015),
  \url{https://doi.org/10.1093/alh/ajv029}

\bibitem{tang_languagespecific_2024a}
Tang, T., Luo, W., Huang, H., Zhang, D., Wang, X., Zhao, X., Wei, F., Wen,
  J.R.: Language-{{Specific Neurons}}: {{The Key}} to {{Multilingual
  Capabilities}} in {{Large Language Models}} (Jun 2024),
  \url{http://arxiv.org/abs/2402.16438}

\bibitem{todorov_assessment_2022a}
Todorov, K., Colavizza, G.: An {{Assessment}} of the {{Impact}} of {{OCR
  Noise}} on {{Language Models}} (Jan 2022),
  \url{http://arxiv.org/abs/2202.00470}

\bibitem{touvron_Llama_2023b}
Touvron, H., Martin, L., Stone, K., Albert, P., Almahairi, A., Babaei, Y.,
  Bashlykov, N., Batra, S., Bhargava, P., Bhosale, S., Bikel, D., Blecher, L.,
  Ferrer, C.C., Chen, M., Cucurull, G., Esiobu, D., Fernandes, J., Fu, J., Fu,
  W., Fuller, B., Gao, C., Goswami, V., Goyal, N., Hartshorn, A., Hosseini, S.,
  Hou, R., Inan, H., Kardas, M., Kerkez, V., Khabsa, M., Kloumann, I., Korenev,
  A., Koura, P.S., Lachaux, M.A., Lavril, T., Lee, J., Liskovich, D., Lu, Y.,
  Mao, Y., Martinet, X., Mihaylov, T., Mishra, P., Molybog, I., Nie, Y.,
  Poulton, A., Reizenstein, J., Rungta, R., Saladi, K., Schelten, A., Silva,
  R., Smith, E.M., Subramanian, R., Tan, X.E., Tang, B., Taylor, R., Williams,
  A., Kuan, J.X., Xu, P., Yan, Z., Zarov, I., Zhang, Y., Fan, A., Kambadur, M.,
  Narang, S., Rodriguez, A., Stojnic, R., Edunov, S., Scialom, T.: Llama 2:
  {{Open Foundation}} and {{Fine-Tuned Chat Models}} (Jul 2023),
  \url{http://arxiv.org/abs/2307.09288}

\bibitem{vanstrien_assessing_2020}
{van Strien}, D., Beelen, K., Ardanuy, M., Hosseini, K., McGillivray, B.,
  Colavizza, G.: Assessing the {{Impact}} of {{OCR Quality}} on {{Downstream
  NLP Tasks}}. In: Proceedings of the 12th {{International Conference}} on
  {{Agents}} and {{Artificial Intelligence}}. pp. 484--496. {SCITEPRESS -
  Science and Technology Publications}, Valletta, Malta (2020),
  \url{http://www.scitepress.org/DigitalLibrary/Link.aspx?doi=10.5220/0009169004840496}

\bibitem{vaswani_attention_2017a}
Vaswani, A., Shazeer, N., Parmar, N., Uszkoreit, J., Jones, L., Gomez, A.N.,
  Kaiser, {\L}., Polosukhin, I.: Attention is all you need. In: Guyon, I.,
  Luxburg, U.V., Bengio, S., Wallach, H., Fergus, R., Vishwanathan, S.,
  Garnett, R. (eds.) Advances in Neural Information Processing Systems.
  vol.~30, pp. 5998--6008. Curran Associates, Inc., Long Beach, California, US
  (2017),
  \url{https://proceedings.neurips.cc/paper/2017/file/3f5ee243547dee91fbd053c1c4a845aa-Paper.pdf}

\bibitem{wang_finding_2022}
Wang, X., Wen, K., Zhang, Z., Hou, L., Liu, Z., Li, J.: Finding {{Skill
  Neurons}} in {{Pre-trained Transformer-based Language Models}}. In: Goldberg,
  Y., Kozareva, Z., Zhang, Y. (eds.) Proceedings of the 2022 {{Conference}} on
  {{Empirical Methods}} in {{Natural Language Processing}}. pp. 11132--11152.
  Association for Computational Linguistics, Abu Dhabi, United Arab Emirates
  (Dec 2022), \url{https://aclanthology.org/2022.emnlp-main.765}

\end{thebibliography}

\appendix

\section{Fine-tuning}
\label{appendix:finetuning}

We fine-tune Llama2 with the low-rank adaptation technique (LoRA) in which we which aim specific parts of the model (in this case, the internal representation up projection layers) by introducing new parameters that are trained during the fine-tuning process \cite{hu_lora_2021}. These projection layers--the up projections--are responsible for compressing and adjusting the model's high-dimensional representations into a form suitable for output predictions.

The code for NER was implemented as follows. We tokenise the input sequence \( S \) and the resulting tokens \( T \) are fed into Llama2 to extract the latent representation \( H \) for sequence classification $T = \text{Tokenizer}(S)$. The tokens are fed into the pretrained model (Llama2) to get the hidden states \( H \) for each token $H_{\text{seq}} = \text{Model}(T)$. For token classification, the hidden state for each token is used directly $H = H_{\text{seq}}$. Then finally, a fully connected layer and Softmax are applied to each token's representation to map it to the label space $\text{logits}_i = \text{Softmax}(W H_i + b)$ where \( H_i \) is the hidden state for the token \( i \). 

We are aware that in the context of NER, removing causal masks would allow the model to utilise bidirectional context, which could be beneficial. However, it has been shown to have no or low impact on NER \cite{li_label_2023c}. The causal masks in the decoder blocks prevent information leakage, as the decoder is only allowed to attend to earlier positions in text generation. However, considering our goal to analyse OCR-sensitive neurons and their influence on NER, it was advantageous to use the model with causal masks intact to maintain the integrity of the sequence information flow.




\section{Improvement Scores per Dataset}\label{appendix:improvement_scores_per_dataset}

\begin{figure*}[ht]
    \centering
    \includegraphics[width=1.\textwidth]{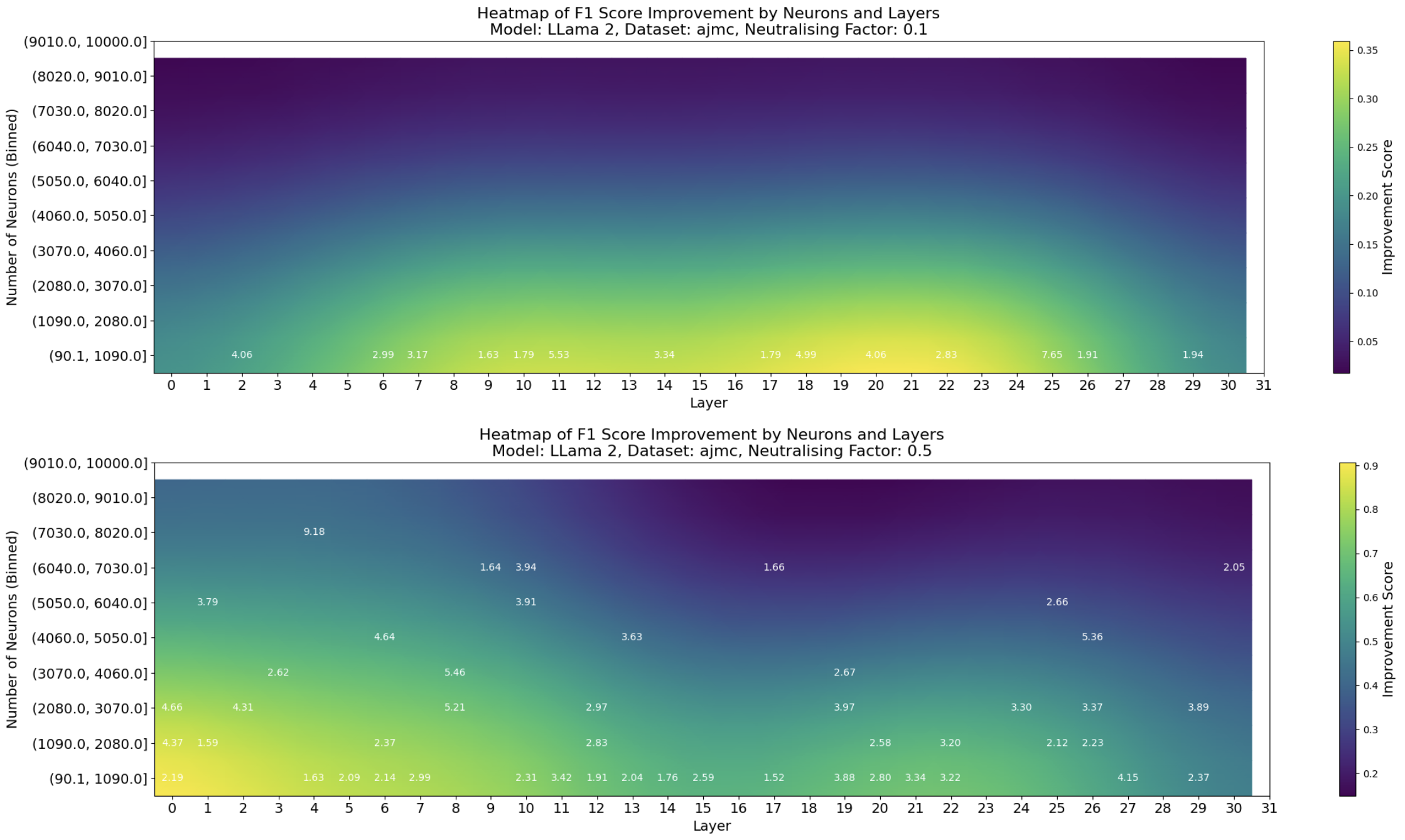}
    \label{fig:improvement_heatmap_ajmc_0.1_0.5_llama2}
    \caption{The improvement scores values in F1 across different layers in Llama2 and neuron bins for the \texttt{ajmc} dataset with a neutralising factor of 0.1 and 0.5.}
\end{figure*}

\begin{figure*}[ht]
    \centering
    \includegraphics[width=1.\textwidth]{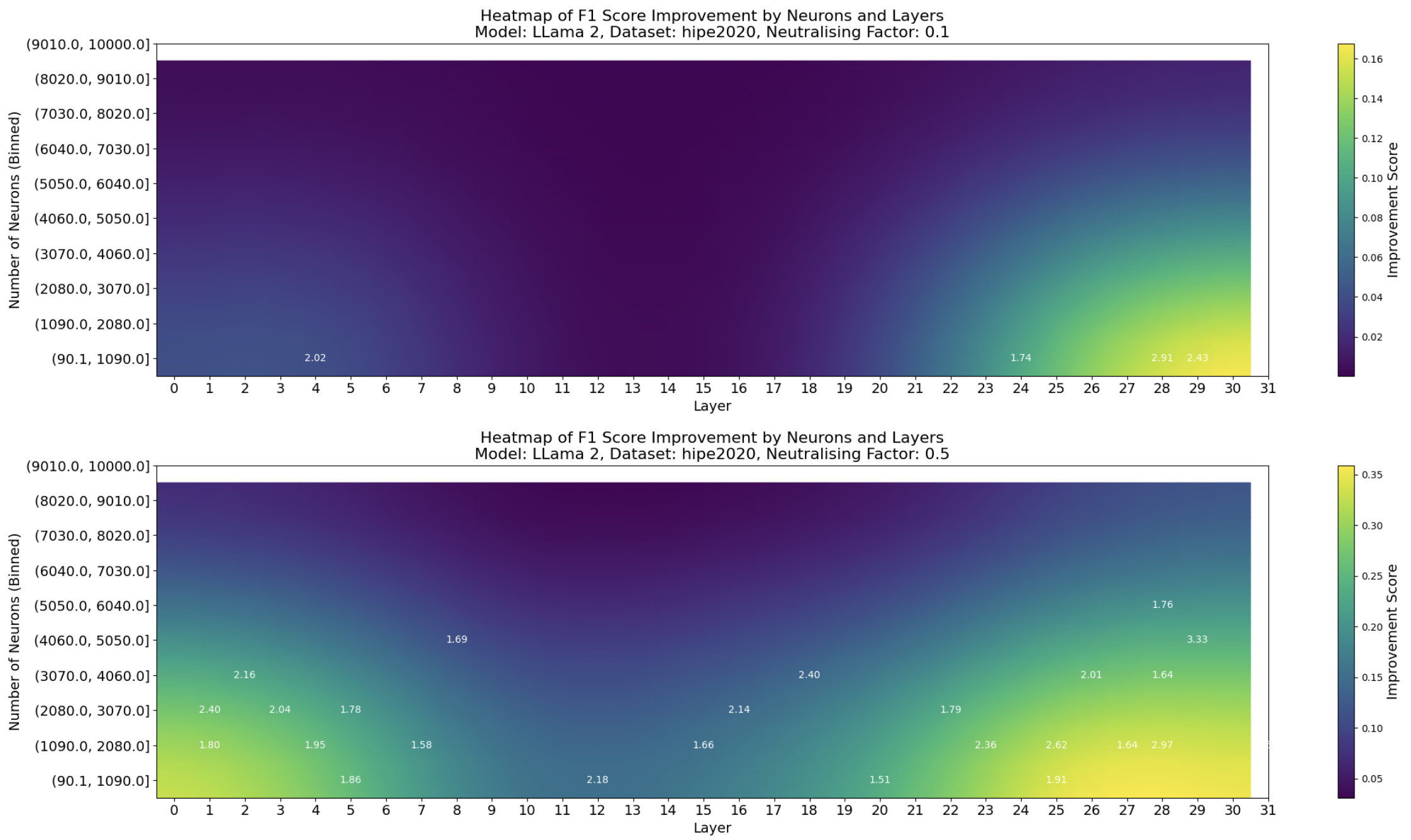}
    \label{fig:improvement_heatmap_hipe2020_0.1_0.5_llama2}
    \caption{The improvement scores values in F1 across different layers in Llama2 and neuron bins for the \texttt{hipe2020} dataset with a neutralising factor of 0.1 and 0.5.}
\end{figure*}

\begin{figure*}[ht]
    \centering
    \includegraphics[width=1.\textwidth]{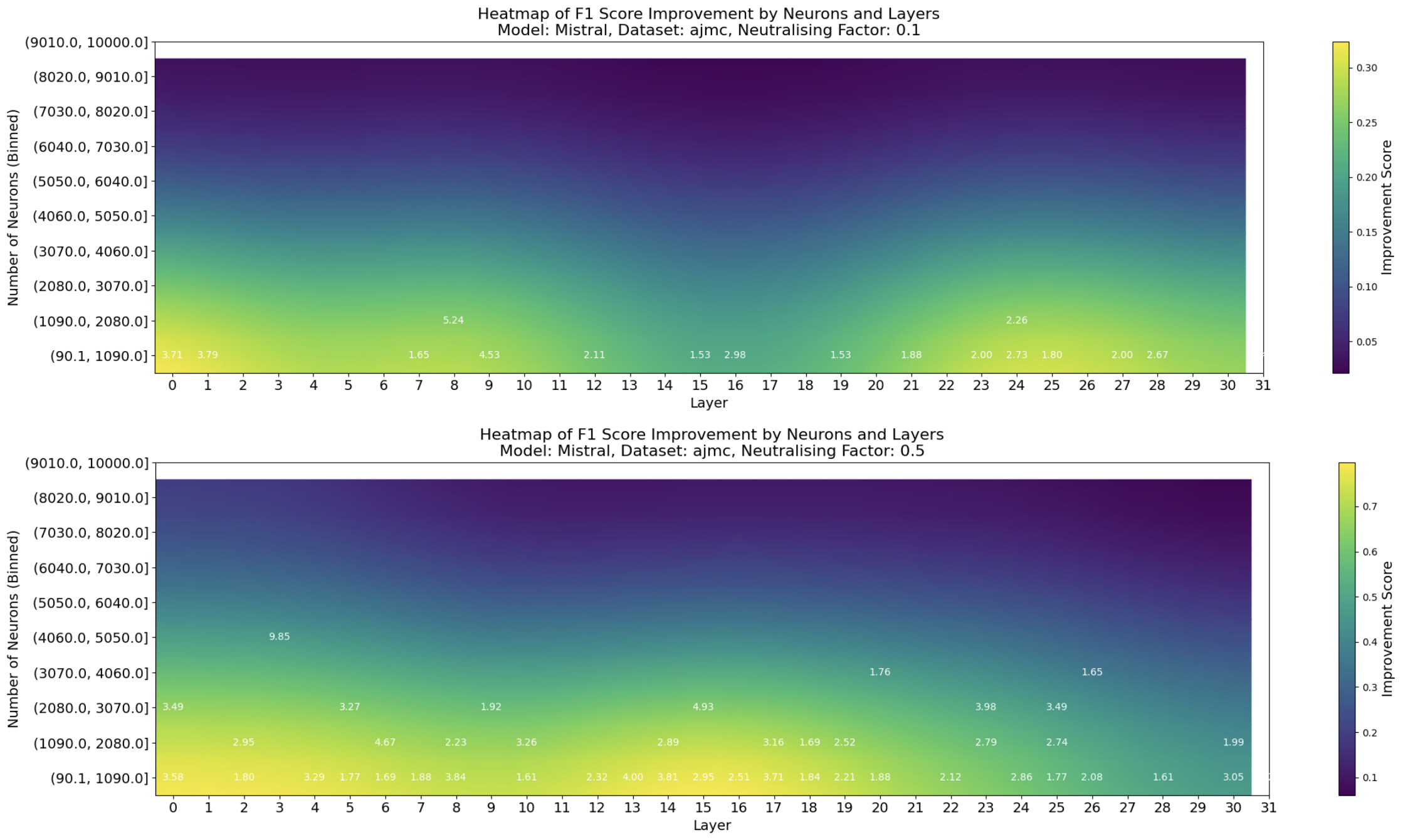}
    \label{fig:improvement_heatmap_ajmc_0.1_0.5_mistral}
    \caption{The improvement scores values in F1 across different layers in Mistral and neuron bins for the \texttt{ajmc} dataset with a neutralising factor of 0.1 and 0.5.}
\end{figure*}

\begin{figure*}[ht]
    \centering
    \includegraphics[width=1.\textwidth]{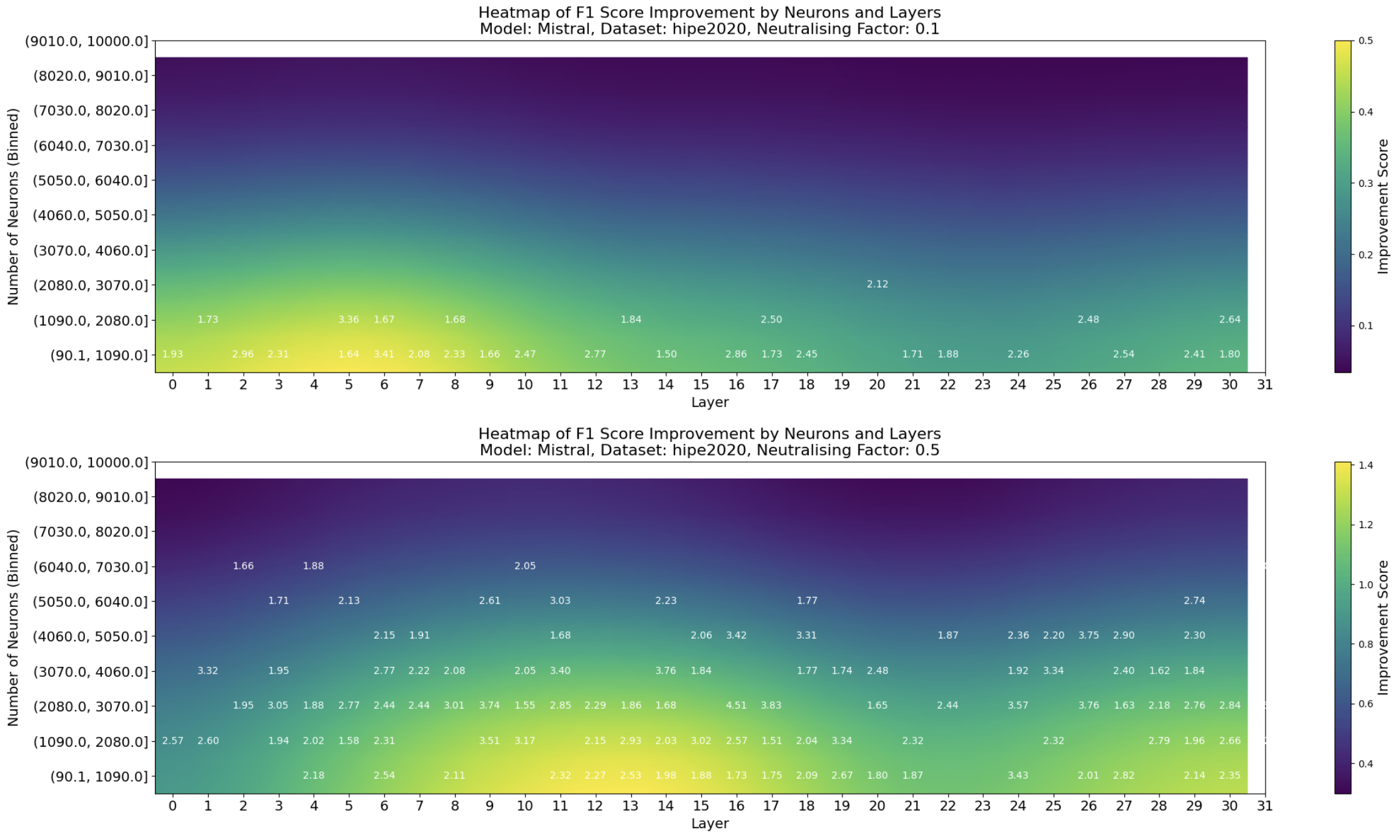}
    \label{fig:improvement_heatmap_hipe2020_0.1_0.5_mistral}
    \caption{The improvement scores values in F1 across different layers in Mistral and neuron bins for the \texttt{hipe2020} dataset with a neutralising factor of 0.1 and 0.5.}
\end{figure*}


\end{document}